%% file: main.tex
\documentclass[11pt]{article}
\usepackage[margin=1in]{geometry}
\usepackage{amsmath}
\usepackage{amssymb}
\usepackage{booktabs}
\usepackage{graphicx}
\usepackage{xcolor}
\usepackage{tikz}
\usepackage{float}
\usepackage{url}
\usepackage{microtype}
\usepackage{fancyhdr}
\usepackage{placeins}
\usepackage{dblfloatfix}
\usepackage{caption}
\usepackage{titlesec}
\usepackage[numbers,sort&compress]{natbib}
\usepackage{orcidlink}
\usepackage{hyperref}   
\hypersetup{colorlinks=true, linkcolor=blue!50!black, citecolor=blue!50!black, urlcolor=blue!50!black}

\graphicspath{{figures/}}
\input{figures/tikz_setup}

\newif\ifshowtodos
\showtodostrue

\title{%
  \includegraphics[width=0.18\textwidth]{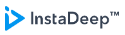}\\[0.7em]
  Advancing Clinical Trial Outcome Prediction with Domain-Specific Large Language Models
}

\author{
  \begin{tabular}{c@{\hspace{3em}}c}
    Burak Bakkar$^{*}$\,\orcidlink{0009-0005-1540-0204} & Hana Moalla$^{*}$\,\orcidlink{0009-0005-0242-7643}\\
    \small \href{mailto:b.bakkar@instadeep.com}{b.bakkar@instadeep.com}
      & \small \href{mailto:h.moalla@instadeep.com}{h.moalla@instadeep.com}
  \end{tabular}\\[2.0em]
  Mohamed Salem Jedidi\,\orcidlink{0009-0009-2544-6096}, Haifa Ben Messaoud\,\orcidlink{0000-0001-5557-8203},
  Nicolas Lopez Carranza\,\orcidlink{0000-0003-2235-5753}\\[0.3em]
  \small InstaDeep\\[0.5em]
  \small $^{*}$Equal contributions.
}
\date{}

\begin{document}
\maketitle

\input{sections/abstract}

\vspace{0.5em}

\vspace{0.5em}

\input{sections/introduction}
\input{sections/related_work}
\clearpage
\input{sections/methods}
\input{sections/results}
\input{sections/discussion}
\input{sections/conclusion}
\input{sections/declarations}

\clearpage
\bibliographystyle{unsrtnat}
\bibliography{references}

\end{document}

%% file: figures/tikz_setup.tex
\usetikzlibrary{positioning, arrows.meta, fit, backgrounds, calc, shapes.misc}
\definecolor{boxline}{HTML}{4D4D4D}
\definecolor{arrowgray}{HTML}{555555}
\definecolor{sharedfill}{HTML}{EDEDED}
\definecolor{evalfill}{HTML}{DDEFD9}
\definecolor{trainfill}{HTML}{F9C9A8}   
\definecolor{trainline}{HTML}{D55E00}
\definecolor{frozenfill}{HTML}{CFE3F2}  
\definecolor{frozenline}{HTML}{0072B2}
\definecolor{probelane}{HTML}{EEF5FB}
\definecolor{ftlane}{HTML}{FDF1E8}

%% file: sections/abstract.tex
\begin{center}
    {\Large\bfseries Abstract}
\end{center}
Clinical trial failures are a major barrier to efficient drug development, incurring substantial costs and delaying patient access to effective therapies. Structured approaches such as the Hierarchical Interaction Network (HINT) predict trial outcomes by explicitly modelling interactions between drugs, diseases and protocols, but make limited use of the free text in trial registrations. We evaluate two domain-specific language models, BioBERT (108M parameters) and BioMistral (7B parameters), for phase-level outcome prediction on the TOP benchmark (12,477 trials), representing each trial as text built from ten ClinicalTrials.gov fields. We compare fine-tuning (full fine-tuning for BioBERT, low-rank adaptation for BioMistral) with probing, in which the pretrained model is frozen and only a lightweight classifier is trained on its embeddings. On the held-out test set of each phase, all four language-model variants achieved higher ROC-AUC (0.613--0.772) and PR-AUC (0.649--0.901) than the published HINT results, corresponding to relative PR-AUC gains of 10--20\%. F1 differences were small and mixed, and no model, including HINT, clearly exceeded the F1 of a trivial classifier that predicts success for every trial. Under the configurations tested, probing performed comparably to fine-tuning: fine-tuned models scored highest on ROC-AUC and PR-AUC in Phases~I and~II, and frozen BioMistral embeddings in Phase~III (ROC-AUC 0.772; PR-AUC 0.901), although most differences between variants were below 0.03 and were not tested for significance. These results suggest that pretrained biomedical representations already encode signal relevant to trial outcome, and that probing is a computationally cheaper alternative to fine-tuning for this task.

%% file: sections/introduction.tex
\section{Introduction}
Clinical trials are the most costly and failure-prone stage of drug development. Roughly 86--90\% of drug candidates that enter clinical testing never receive regulatory approval~\cite{sun2022why90,wong2019estimation}, and attrition remains high in the later, more expensive phases~\cite{fogel2018factors,wong2019estimation}. Each failed trial represents a large research and development loss and delays patient access to effective therapies~\cite{fogel2018factors}. Reliable estimates of a trial's probability of success before it starts could help sponsors prioritise portfolios, refine protocols and allocate resources.

Early computational approaches relied on structured data and manual feature engineering. Drug, disease and trial characteristics were encoded as handcrafted variables, with drug properties often represented through simplified molecular-input line-entry system (SMILES) strings or chemical descriptors, and then passed to classical machine-learning models~\cite{gayvert2016data,lo2019machine,hong2020ensemble}. These representations captured basic chemical and design information but not the interactions among drugs, diseases and patient populations, and their performance often failed to generalise across therapeutic areas.

The Hierarchical Interaction Network (HINT) addressed this by modelling these interactions explicitly~\cite{fu2020hint}. HINT combines molecular, disease-ontology, pharmacokinetic and eligibility-criteria encoders with a graph neural network, and was introduced together with the Trial Outcome Prediction (TOP) benchmark for phase-level trial outcome prediction. HINT's only free-text input, however, is the eligibility criteria; other free text in trial registrations, such as brief summaries, intervention descriptions and outcome definitions, is not used.

Domain-specific pretrained language models offer a way to exploit this text. Models such as BioBERT~\cite{lee2020biobert} and PubMedBERT~\cite{gu2021domain}, and more recent large models such as BioMistral~\cite{labrak2024biomistral}, learn contextual representations of biomedical terminology from large literature corpora, and have performed well on a wide range of biomedical language tasks~\cite{lee2020biobert,gu2021domain,labrak2024biomistral}. Recent work has begun to apply language models to trial outcome prediction, representing trials partly or wholly as text~\cite{gao2024lint,zheng2025lifted}. It remains unclear, however, how much of the predictive signal is already present in pretrained representations and how much must be learned through task-specific fine-tuning, and how compact encoder models compare with larger decoder models on this task.

In this work, we follow this text-based approach: we represent each trial as text assembled from ten ClinicalTrials.gov fields that describe its summary, condition, intervention, design and primary outcome, and evaluate BioBERT and BioMistral on the TOP benchmark under two training strategies. Our contributions are threefold. First, we provide a systematic comparison of fine-tuning and probing for an encoder model (BioBERT, 108M parameters) and a decoder model (BioMistral, 7B parameters), evaluated separately for Phase~I, II and III trials and compared against published HINT results. Second, we introduce a prevalence-aware evaluation for this benchmark: we report a trivial always-success reference and a prevalence-normalised area under the precision--recall curve (PR-AUC), and show that F1 at a fixed threshold does not separate the models from this reference. Third, we show that, under the configurations tested, frozen biomedical representations with a small classifier perform comparably to fine-tuned models.

%% file: sections/related_work.tex
\section{Related work}

\textbf{Feature-based prediction of trial success.}
Early studies predicted trial or drug-programme success from engineered features. Gayvert et al.\ combined chemical properties of drugs with features of their molecular targets~\cite{gayvert2016data}, Lo et al.\ used statistical imputation and machine-learning models on drug and trial characteristics to predict approvals~\cite{lo2019machine}, and Hong et al.\ applied an ensemble of least-squares support vector regressors~\cite{hong2020ensemble}. A recent scoping review summarises the wider use of artificial intelligence for clinical trial risk assessment~\cite{teodoro2025scoping}.

\textbf{HINT and the TOP benchmark.}
HINT introduced the TOP benchmark and a hierarchical interaction network that integrates drug molecules, disease codes, pharmacokinetic properties and eligibility criteria~\cite{fu2020hint}. Several methods build on this benchmark. SPOT (sequential predictive modelling of clinical trial outcome) clusters trials into topics, orders them over time and applies meta-learning, reporting PR-AUC gains over earlier methods in all three phases~\cite{wang2023spot}. Lu et al.\ added selective classification to HINT, allowing the model to abstain on low-confidence trials, and reported a Phase~III PR-AUC of 0.902 on the trials it retained~\cite{lu2024uncertainty}; because that score is computed on a subset of the test set, it is not directly comparable to full-test-set results.

\textbf{Language models for trial outcome prediction.}
The Language Interaction Network (LINT) predicts trial outcomes primarily from free-text trial descriptions and reported areas under the receiver operating characteristic curve (ROC-AUC) of 0.770, 0.740 and 0.748 for Phases~I--III on trials of biologic interventions~\cite{gao2024lint}. LIFTED converts each data modality into natural-language descriptions with a large language model and combines the resulting representations through mixture-of-experts modules~\cite{zheng2025lifted}. Language models have also been applied to related trial-level tasks such as predicting enrollment success~\cite{yue2024trialenroll}.

\textbf{Biomedical language models and probing.}
Domain-specific pretraining on biomedical literature improves performance on biomedical language tasks: BioBERT continues the pretraining of BERT (Bidirectional Encoder Representations from Transformers) on PubMed abstracts and, in some versions, PubMed Central full-text articles~\cite{lee2020biobert,devlin2019bert}, and PubMedBERT is pretrained on PubMed text from scratch~\cite{gu2021domain}. Large models such as Med-PaLM encode substantial clinical knowledge~\cite{singhal2023large}, and BioMistral adapts the instruction-tuned Mistral~7B model (Mistral-7B-Instruct-v0.1)~\cite{jiang2023mistral} through further pretraining on PubMed Central~\cite{labrak2024biomistral}. Pretraining has also been applied to structured electronic health records~\cite{rasmy2021medbert}. Probing classifiers, which are trained on frozen representations, are a standard tool for measuring what information a pretrained model already encodes~\cite{belinkov2022probing}.

\textbf{Positioning.}
Our study is closest to LINT, which also relies on trial text. It differs in three respects: we evaluate on the full TOP benchmark rather than on biologic interventions only; we compare an encoder model with a 7B decoder model; and we contrast fine-tuning with probing to separate the contribution of pretrained representations from that of task-specific training. Because the cited studies use different subsets, splits or evaluation protocols, their reported scores should not be compared directly with ours.

%% file: sections/methods.tex
\section{Methods}
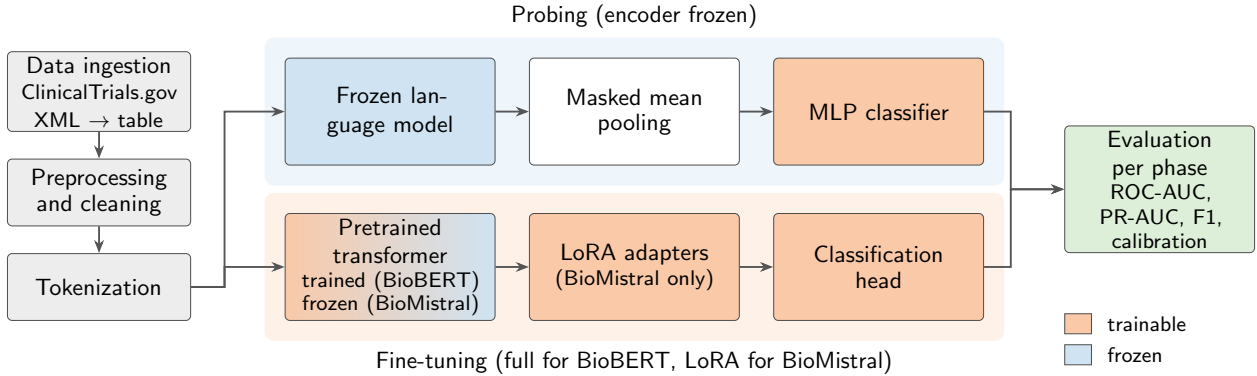
\begin{figure}[H]
  \centering
  \input{figures/fig_pipeline}
  \caption{Clinical trial outcome prediction pipeline. Trial records are parsed from ClinicalTrials.gov, cleaned and tokenized, and then used either to train a classifier on frozen language-model embeddings (probing, top) or to fine-tune the language model end to end (bottom). Orange components are trained; blue components are frozen. Abbreviations: LoRA, low-rank adaptation; MLP, multi-layer perceptron; PR-AUC, area under the precision--recall curve; ROC-AUC, area under the receiver operating characteristic curve; XML, Extensible Markup Language.}
  \label{fig:pipeline}
\end{figure}

Our pipeline covers data collection, preprocessing, text representation, model training and evaluation for predicting whether a clinical trial succeeds or fails (Fig.~\ref{fig:pipeline}).

\subsection{Task definition}
The task is binary classification. For a set of trials $X=\{x_1,\ldots,x_N\}$ with outcomes $Y=\{y_1,\ldots,y_N\}$, $y_n=1$ denotes success and $y_n=0$ failure. Following the HINT definition~\cite{fu2020hint}, a trial is labelled successful when it was completed and met its primary endpoint.

\subsection{Dataset}\label{sec:dataset}
We used the TOP clinical trial outcome prediction benchmark of Fu et al.~\cite{fu2020hint,futianfan_top_data} for trial selection, labels and data splits. TOP contains interventional trials of small-molecule drugs, whose binary outcome labels were manually curated by an internal IQVIA team~\cite{fu2020hint}. The splits are defined by time: training and validation trials were completed before 1 January 2014, and test trials started after that date. The benchmark is released for non-commercial use only~\cite{futianfan_top_data}. We used the data files of its public repository at commit \texttt{d1f5c4a} (30 December 2021), the last commit that modified them. They contain 1,788 Phase~I, 6,105 Phase~II and 4,584 Phase~III trials (12,477 in total), each with predefined training, validation and test splits (Table~\ref{tab:dataset-stats}). This differs slightly from the version described in the HINT publication, which has 12,465 trials and test sets of 627, 1,653 and 1,140 trials~\cite{fu2020hint}. Success rates differ between splits: the Phase~II and Phase~III test sets have higher success rates than the corresponding training sets (55.5\% versus 47.8\%, and 75.0\% versus 65.3\%), whereas Phase~I is similar (55.3\% versus 56.7\%).

\begin{table}[H]
\centering
\small
\setlength{\tabcolsep}{5pt}
\caption{Dataset statistics by phase: number of trials in the training, validation and test splits, number of unique diseases, outcome distribution over all splits, and success rate in each split. Counts and rates were computed from the public TOP data files (commit \texttt{d1f5c4a}).}
\label{tab:dataset-stats}
\begin{tabular}{lccccc}
\toprule
Phase & Trials & Diseases & Success & Failure & Success rate (\%) \\
      & (train / val / test) & & & & (train / val / test) \\
\midrule
Phase I   & 1,044 / 117 / 627   & 1,392 & 1,006 (56.3\%) & 782 (43.7\%)   & 56.7 / 57.3 / 55.3 \\
Phase II  & 4,005 / 446 / 1,654 & 2,824 & 3,040 (49.8\%) & 3,065 (50.2\%) & 47.8 / 46.6 / 55.5 \\
Phase III & 3,094 / 344 / 1,146 & 1,619 & 3,109 (67.8\%) & 1,475 (32.2\%) & 65.3 / 66.6 / 75.0 \\
\bottomrule
\end{tabular}
\end{table}

\subsubsection{Data curation}
\begin{figure}[H]
  \centering
  \includegraphics[width=0.92\textwidth]{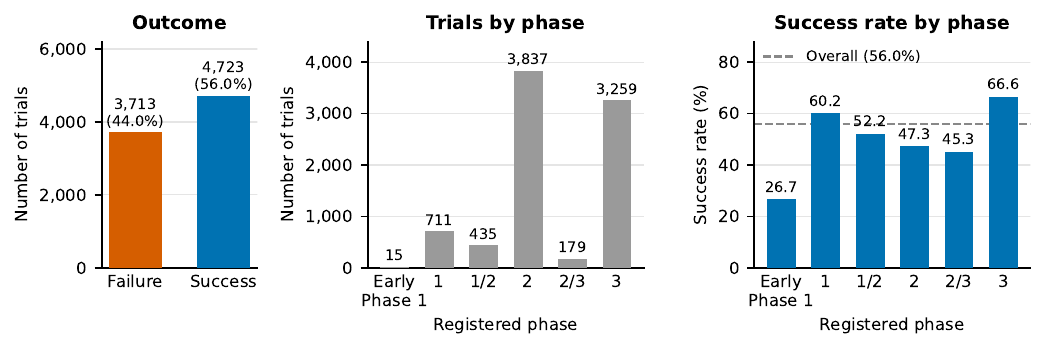}
  \caption{Composition of the merged training and validation data (8,436 trials): outcome distribution (left), number of trials by registered phase (middle), and success rate by registered phase (right). The dashed line marks the overall success rate.}
  \label{fig:dataset-composition}
\end{figure}

The benchmark was downloaded from its GitHub repository~\cite{futianfan_top_data} and provides separate training, validation and test files for each phase. Because the per-phase training sets are small, we merged the three training files into a single training set and the three validation files into a single validation set, and kept the three test sets separate. Trials registered as combined phases (Phase~I/II or Phase~II/III) appear in more than one phase file; removing duplicate records reduced the merged training set from 8,143 to 7,641 unique trials and the merged validation set from 907 to 901 unique trials. We then removed the 106 validation trials that also appeared in the training set, leaving 795 validation trials. Fig.~\ref{fig:dataset-composition} summarises the composition of the cleaned training and validation data. No test trial appears in the merged training or validation sets.

\subsubsection{Feature extraction from ClinicalTrials.gov}
The TOP files do not contain all the fields we required, so we retrieved each trial's full registry record from ClinicalTrials.gov~\cite{clinicaltrialsgov} using its National Clinical Trial (NCT) identifier, a unique code of the form ``NCT'' followed by eight digits (for example, NCT00000419). Records were downloaded as Extensible Markup Language (XML) files on 12 September 2024 and parsed into a table of 253 variables. All 12,477 trials in the benchmark were matched. These records reflect each registry entry as of the download date rather than as originally registered; we return to this point in Section~\ref{sec:limitations}.

\subsection{Trial representation}
\begin{figure}[H]
  \centering
  \input{figures/fig_input_representation}
  \caption{Input representation. For fine-tuning, the ten fields are concatenated into one sequence that is passed through the trainable language model. For probing, each field is encoded separately by the frozen model, token embeddings are mean-pooled per field and averaged across fields, and the result is passed to an MLP classifier. Abbreviations: LoRA, low-rank adaptation; MLP, multi-layer perceptron.}
  \label{fig:input-representation}
\end{figure}
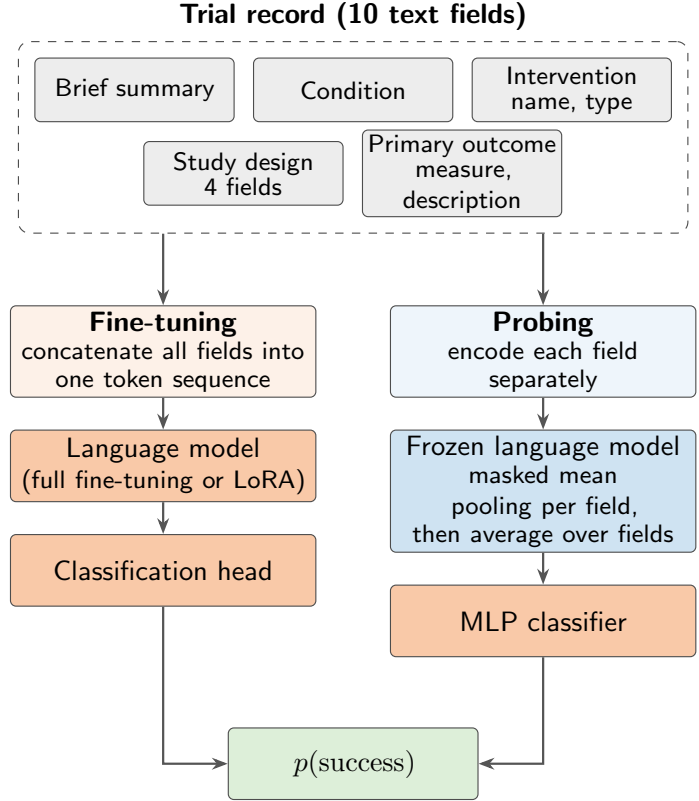

Each trial $x_n$ is represented by ten fields that describe its design and intent. Fig.~\ref{fig:input-representation} shows how the two training strategies use them.

The ten fields are:
\begingroup\raggedright
\begin{itemize}
  \item[-] brief summary (\nolinkurl{brief_summary/textblock});
  \item[-] condition (\nolinkurl{condition});
  \item[-] intervention name and type (\nolinkurl{intervention/intervention_name}, \nolinkurl{intervention/intervention_type});
  \item[-] study design: allocation, intervention model, masking and primary purpose (\nolinkurl{study_design_info/allocation}, \nolinkurl{intervention_model}, \nolinkurl{masking}, \nolinkurl{primary_purpose});
  \item[-] primary outcome measure and description (\nolinkurl{primary_outcome/measure}, \nolinkurl{primary_outcome/description}).
\end{itemize}
\endgroup
The phase of the trial is not among these fields. Because a single model is trained on trials from all phases, phase information is available to the model only where it is mentioned in the text itself.

The two training strategies use these fields differently (Fig.~\ref{fig:input-representation}). For fine-tuning, the fields are concatenated into a single text sequence in the following order: brief summary, condition, intervention, study design and primary outcome. For probing, each field is encoded separately and the resulting field embeddings are averaged (Section~\ref{sec:probing}). For fine-tuning, the concatenated input was truncated to 512 tokens for both models. For probing, each field was truncated separately, to 512 tokens for BioBERT and 2,048 tokens for BioMistral, the sequence length used in BioMistral's pretraining.

\subsection{Language models}
\textbf{BioBERT} (108M parameters)~\cite{lee2020biobert} is initialised from the cased BERT-base model~\cite{devlin2019bert} and further pretrained on biomedical text. We used version 1.1 (\nolinkurl{dmis-lab/biobert-v1.1}), which was pretrained on PubMed abstracts for 1M steps. \textbf{BioMistral} (7B parameters)~\cite{labrak2024biomistral} is obtained by further pretraining Mistral-7B-Instruct-v0.1, the instruction-tuned version of Mistral~7B~\cite{jiang2023mistral}, on PubMed Central open-access articles with a sequence length of 2,048 tokens. We used the \nolinkurl{BioMistral/BioMistral-7B} checkpoint. BioBERT produces 768-dimensional hidden states and BioMistral 4,096-dimensional hidden states.

\subsection{Fine-tuning}
\begin{figure}[H]
  \centering
  \input{figures/fig_training_strategies}
  \caption{Training strategies. (a) Full fine-tuning updates all transformer weights and the classification head. (b) LoRA keeps the pretrained weights frozen and trains low-rank adapters ($\Delta W = (\alpha/r)\,BA$) inserted into the attention projections, together with the classification head. (c) Probing uses the frozen model only as a feature extractor; token embeddings are mean-pooled and only the MLP classifier is trained. Abbreviations: LoRA, low-rank adaptation; MLP, multi-layer perceptron.}
  \label{fig:training-strategies}
\end{figure}
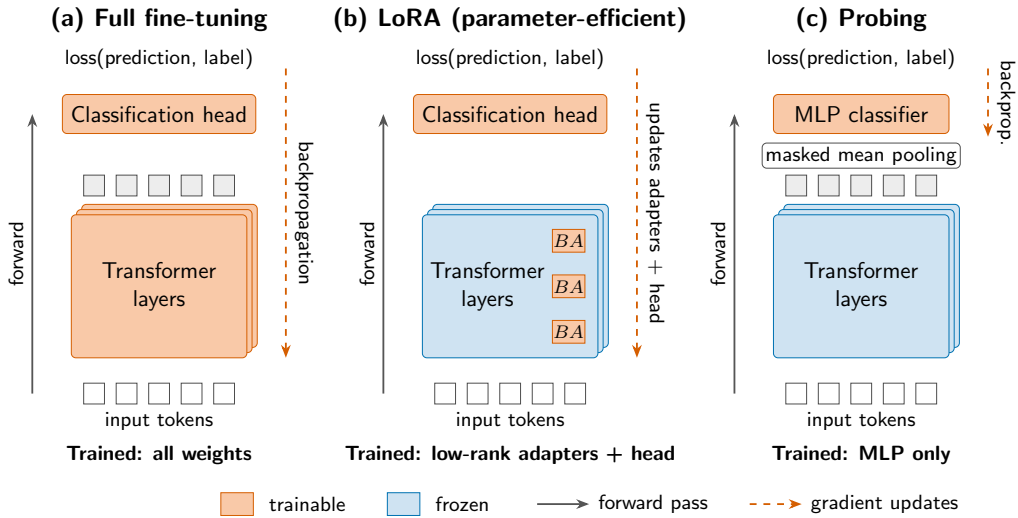

We fine-tuned BioBERT with full fine-tuning and BioMistral with low-rank adaptation (LoRA)~\cite{hu2021lora}, a parameter-efficient fine-tuning method (Fig.~\ref{fig:training-strategies}a,b).

In full fine-tuning, all pretrained weights are updated by gradient descent on the downstream task. This gives the model maximum flexibility to adapt its representations but requires storing gradients and optimiser states for every parameter. LoRA instead keeps the pretrained weights frozen and adds a trainable low-rank update $\Delta W = (\alpha/r)\,BA$ to selected weight matrices, where $B\in\mathbb{R}^{d\times r}$, $A\in\mathbb{R}^{r\times k}$, $r \ll \min(d,k)$ and $\alpha$ is a scaling factor. Only $A$, $B$ and the classification head are trained, which greatly reduces memory and compute.

\subsubsection{Hyperparameter settings}
We evaluated several configurations, including mixed-precision training and different learning-rate schedulers (OneCycleLR, ReduceLROnPlateau and others). All models were trained with the Adam optimiser~\cite{kingma2015adam} without weight decay, using the unweighted binary cross-entropy loss. Configurations were compared on the validation set, and we report the best-performing configuration for each model.

\textbf{BioBERT full fine-tuning.} Up to five epochs, mini-batch size 32, float16 mixed precision, Adam optimiser with learning rate $3\times10^{-5}$ (no weight decay), and early stopping on the validation loss (patience 2, minimum improvement $\delta=10^{-4}$). No learning-rate scheduler was used for the reported run.

\textbf{BioMistral LoRA fine-tuning.} LoRA with rank $r=32$, scaling factor $\alpha=32$ and dropout 0.01, applied to the query and value projections (\nolinkurl{q_proj}, \nolinkurl{v_proj}) for sequence classification. This yields 13,631,488 trainable adapter parameters (0.19\% of all parameters). The model was trained for up to five epochs with mini-batch size 2 in float16 mixed precision, using the Adam optimiser with a learning rate of $3\times10^{-5}$ and no learning-rate scheduler. Early stopping was applied to the validation loss with a patience of 2 and a minimum improvement of $\delta=10^{-4}$.

\subsection{Probing}\label{sec:probing}
Probing measures how much predictive signal is already present in pretrained representations, independently of task-specific adaptation~\cite{belinkov2022probing}. The weights of the pretrained model are kept frozen and only a lightweight classifier is trained (Fig.~\ref{fig:training-strategies}c). Because the frozen model is run only once for each field of each trial to compute embeddings, which are then stored, probing is also much cheaper to train than fine-tuning.

For a text field tokenized into $n$ tokens, the frozen model produces hidden states $H=[\mathbf{h}_1,\ldots,\mathbf{h}_n]\in\mathbb{R}^{n\times d}$, where $d=768$ for BioBERT and $d=4096$ for BioMistral. We obtain a fixed-size field embedding by masked mean pooling,
\begin{equation}
\mathbf{e}_f=\frac{\sum_{i=1}^{n} m_i\,\mathbf{h}_i}{\sum_{i=1}^{n} m_i},
\label{eq:meanpool}
\end{equation}
where $m_i\in\{0,1\}$ is the attention mask that excludes padding tokens. The hidden states are taken from the final layer. The trial embedding $\mathbf{e}$ is the average of the field embeddings $\mathbf{e}_f$.

The trial embedding is passed to a multi-layer perceptron (MLP) with $L$ hidden layers. Starting from $\mathbf{h}^{(0)}=\mathbf{e}$, each hidden layer applies a linear transformation, a rectified linear unit (ReLU) activation and dropout:
\begin{equation}
\mathbf{h}^{(l)}=\operatorname{Dropout}\big(\operatorname{ReLU}(\mathbf{W}^{(l)}\mathbf{h}^{(l-1)}+\mathbf{b}^{(l)})\big),
\end{equation}
for $l=1,\ldots,L$, followed by an output layer that produces a single logit $\hat{y}=\mathbf{w}^{\top}\mathbf{h}^{(L)}+b$. The predicted probability of success is $p=\sigma(\hat{y})$, and the classifier is trained with the binary cross-entropy loss
\begin{equation}
\mathcal{L}=-\frac{1}{N}\sum_{i=1}^{N}\big[y_i\log\sigma(\hat{y}_i)+(1-y_i)\log\big(1-\sigma(\hat{y}_i)\big)\big].
\end{equation}
Text representations were taken from the transformer's final hidden layer. Fine-tuned BioBERT used the classification (\texttt{[CLS]}) token, and both probing models (BioBERT and BioMistral) used attention-masked mean pooling over tokens (Eq.~\ref{eq:meanpool}). All MLP hidden layers contained 512 units and used ReLU activation with dropout 0.2. Both probing classifiers used $L=4$ hidden layers, giving 1,182,209 trainable parameters for BioBERT (input size 768) and 2,886,145 for BioMistral (input size 4,096).

\subsubsection{Probing hyperparameter settings}
Probing classifiers were trained with the Adam optimiser and float16 mixed precision. We compared several learning-rate schedules (OneCycleLR, ReduceLROnPlateau, StepLR and others) and used early stopping on the validation loss.

\textbf{BioBERT probing.} Batch size 64, base learning rate $4\times10^{-5}$, dropout 0.2, up to 20 epochs, and no learning-rate scheduler. Training stopped early if the validation loss did not improve by $\delta=10^{-4}$ for three consecutive epochs.

\textbf{BioMistral probing.} Batch size 64, base learning rate $4\times10^{-5}$, OneCycleLR with a maximum learning rate of $10^{-3}$ updated per batch, dropout 0.2, up to 20 epochs with the same early-stopping rule.

\subsection{Baselines}
\textbf{HINT.} We compare against HINT~\cite{fu2020hint}, which combines a drug-molecule encoder based on a message-passing neural network~\cite{huang2020deeppurpose}, a disease-ontology encoder based on the graph-based attention model (GRAM)~\cite{choi2017gram}, a BERT-based eligibility-criteria encoder and a pharmacokinetics encoder, and models their interactions with a graph neural network. HINT results are taken from Table~2 of the original publication, where they are reported as means over 30 independent runs. A separate HINT model was trained for each phase, on the earlier version of the TOP data described in Section~\ref{sec:dataset} (test sets of 627, 1,653 and 1,140 trials). HINT was not retrained in our pooled-phase setting or on our exact test sets, and it uses different input modalities, so the comparison is not fully controlled.

\textbf{Trivial reference.} As a reference point, we also report a classifier that predicts success for every trial. With test-set success rate $\pi$, its F1 is $2\pi/(1+\pi)$, its PR-AUC equals $\pi$ and its ROC-AUC is 0.5.

\subsection{Evaluation metrics}
Models were evaluated separately on the test set of each phase using three metrics:
\begin{itemize}
  \item[-] \textbf{F1:} the harmonic mean of precision and recall for the success class, computed at a fixed decision threshold of 0.5.
  \item[-] \textbf{ROC-AUC:} the area under the receiver operating characteristic curve, which summarises the trade-off between true- and false-positive rates across all thresholds and equals 0.5 for a random classifier.
  \item[-] \textbf{PR-AUC:} the area under the precision--recall curve, computed using average precision, which summarises precision across recall levels. Its value for a random classifier equals the proportion of successful trials, so it should be interpreted relative to the test-set success rate in Table~\ref{tab:dataset-stats}.
\end{itemize}
To compare PR-AUC across phases with different success rates, we also report prevalence-normalised PR-AUC, $(\text{PR-AUC}-\pi)/(1-\pi)$, where $\pi$ is the test-set success rate; it is 0 for a random classifier and 1 for a perfect one. We also report calibration curves, which compare the mean predicted probability with the observed success rate within bins of predicted probability.

%% file: figures/fig_pipeline.tex
\resizebox{\textwidth}{!}{%
\begin{tikzpicture}[
  font=\sffamily\small,
  node distance=4mm and 5mm,
  box/.style={draw=boxline, fill=white, rounded corners=2pt, align=center,
              minimum height=10mm, text width=26mm, inner sep=2pt},
  sbox/.style={box, text width=30mm, minimum height=16mm},
  arr/.style={-{Stealth[length=2mm]}, thick, draw=arrowgray},
  lane/.style={rounded corners=4pt, inner sep=3mm},
]
\node[box, fill=sharedfill] (ingest) {Data ingestion\\[-1pt]{\footnotesize ClinicalTrials.gov XML $\rightarrow$ table}};
\node[box, fill=sharedfill, below=of ingest] (prep) {Preprocessing\\[-1pt]and cleaning};
\node[box, fill=sharedfill, below=of prep] (tok) {Tokenization};
\draw[arr] (ingest) -- (prep);
\draw[arr] (prep) -- (tok);

\node[sbox, right=14mm of ingest, yshift=-3mm, fill=frozenfill] (p1) {Frozen language model};
\node[sbox, right=of p1] (p2) {Masked mean\\[-1pt]pooling};
\node[sbox, right=of p2, fill=trainfill] (p3) {MLP classifier};
\node[sbox, right=14mm of tok, yshift=3mm, left color=trainfill, right color=frozenfill] (f1) {Pretrained transformer\\[-1pt]{\footnotesize trained (BioBERT)\\[-2pt]frozen (BioMistral)}};
\node[sbox, right=of f1, fill=trainfill] (f2) {LoRA adapters\\[-1pt]{\footnotesize (BioMistral only)}};
\node[sbox, right=of f2, fill=trainfill] (f3) {Classification\\[-1pt]head};

\begin{scope}[on background layer]
  \node[lane, fill=probelane, fit=(p1)(p3), label={[font=\sffamily\small]above:Probing (encoder frozen)}] (plane) {};
  \node[lane, fill=ftlane, fit=(f1)(f3), label={[font=\sffamily\small]below:Fine-tuning (full for BioBERT, LoRA for BioMistral)}] (flane) {};
\end{scope}

\draw[arr] (p1) -- (p2); \draw[arr] (p2) -- (p3);
\draw[arr] (f1) -- (f2); \draw[arr] (f2) -- (f3);
\draw[arr] (tok.east) -- ++(5mm,0) |- (p1.west);
\draw[arr] (tok.east) -- ++(5mm,0) |- (f1.west);

\node[box, fill=evalfill, right=12mm of $(p3.east)!0.5!(f3.east)$, text width=27mm] (eval)
  {Evaluation per phase\\[-1pt]{\footnotesize ROC-AUC, PR-AUC, F1,\\[-2pt] calibration}};
\draw[arr] (p3.east) -- ++(4mm,0) |- (eval.west);
\draw[arr] (f3.east) -- ++(4mm,0) |- (eval.west);

\node[draw=boxline, fill=trainfill, minimum width=4mm, minimum height=3mm, inner sep=0pt,
      below=9mm of eval.south west, anchor=north west] (lg) {};
\node[right=1mm of lg, font=\sffamily\footnotesize] (lgt) {trainable};
\node[draw=boxline, fill=frozenfill, minimum width=4mm, minimum height=3mm, inner sep=0pt,
      below=1.5mm of lg] (lg2) {};
\node[right=1mm of lg2, font=\sffamily\footnotesize] {frozen};
\end{tikzpicture}}

%% file: figures/fig_input_representation.tex
\resizebox{0.55\textwidth}{!}{%
\begin{tikzpicture}[
  font=\sffamily\small,
  field/.style={draw=boxline, fill=sharedfill, rounded corners=2pt, align=center,
                minimum height=8mm, text width=24mm, inner sep=1.5pt, font=\sffamily\footnotesize},
  box/.style={draw=boxline, fill=white, rounded corners=2pt, align=center,
              minimum height=9mm, text width=37mm, inner sep=2pt},
  arr/.style={-{Stealth[length=2mm]}, thick, draw=arrowgray},
]
\node[field] (s)  at (0,0)    {Brief summary};
\node[field] (c)  at (2.75,0) {Condition};
\node[field] (i)  at (5.5,0)  {Intervention\\[-2pt]name, type};
\node[field] (d)  at (1.37,-1.05) {Study design\\[-2pt]4 fields};
\node[field] (o)  at (4.12,-1.05) {Primary outcome\\[-2pt]measure, description};
\begin{scope}[on background layer]
  \node[draw=boxline, dashed, rounded corners=4pt, inner sep=2mm, fit=(s)(c)(i)(d)(o),
        label={[font=\sffamily\small\bfseries]above:Trial record (10 text fields)}] (rec) {};
\end{scope}

\node[box, fill=ftlane, below=9mm of rec.south west, anchor=north west, xshift=-1mm] (ft1)
  {\textbf{Fine-tuning}\\[-1pt]{\footnotesize concatenate all fields into\\[-2pt] one token sequence}};
\node[box, below=4mm of ft1, left color=trainfill, right color=trainfill] (ft2)
  {Language model\\[-1pt]{\footnotesize (full fine-tuning or LoRA)}};
\node[box, below=4mm of ft2, fill=trainfill] (ft3) {Classification head};

\node[box, fill=probelane, below=9mm of rec.south east, anchor=north east, xshift=1mm] (pr1)
  {\textbf{Probing}\\[-1pt]{\footnotesize encode each field\\[-2pt] separately}};
\node[box, below=4mm of pr1, fill=frozenfill] (pr2)
  {Frozen language model\\[-1pt]{\footnotesize masked mean pooling per field,\\[-2pt] then average over fields}};
\node[box, below=4mm of pr2, fill=trainfill] (pr3) {MLP classifier};

\draw[arr] (rec.south -| ft1.north) -- (ft1.north);
\draw[arr] (rec.south -| pr1.north) -- (pr1.north);
\draw[arr] (ft1) -- (ft2); \draw[arr] (ft2) -- (ft3);
\draw[arr] (pr1) -- (pr2); \draw[arr] (pr2) -- (pr3);

\node[box, fill=evalfill, below=12mm of $(ft3.south)!0.5!(pr3.south)$, text width=30mm] (out)
  {$p(\mathrm{success})$};
\draw[arr] (ft3.south) |- (out.west);
\draw[arr] (pr3.south) |- (out.east);
\end{tikzpicture}}

%% file: figures/fig_training_strategies.tex
\resizebox{0.82\textwidth}{!}{%
\begin{tikzpicture}[
  font=\sffamily\small,
  tok/.style={draw=boxline, fill=white, minimum size=3.2mm, inner sep=0pt},
  hid/.style={draw=boxline, fill=sharedfill, minimum size=3.2mm, inner sep=0pt},
  head/.style={draw=trainline, fill=trainfill, rounded corners=2pt, minimum width=27mm,
               minimum height=6mm, align=center},
  fwd/.style={-{Stealth[length=2mm]}, thick, draw=arrowgray},
  bwd/.style={-{Stealth[length=2mm]}, thick, dashed, draw=trainline},
  small/.style={font=\sffamily\footnotesize},
]
\newcommand{\stack}[2]{
  \foreach \o in {0.16,0.08} {
    \draw[fill=#1, draw=#2, rounded corners=2pt] (-1.35+\o,0.55+\o) rectangle (1.35+\o,2.75+\o);}
  \draw[fill=#1, draw=#2, rounded corners=2pt] (-1.35,0.55) rectangle (1.35,2.75);}
\newcommand{\tokens}{%
  \foreach \x in {-1.0,-0.5,0,0.5,1.0} {\node[tok] at (\x,0) {};}
  \node[small] at (0,-0.42) {input tokens};}
\newcommand{\hidden}{%
  \foreach \x in {-1.0,-0.5,0,0.5,1.0} {\node[hid] at (\x,3.2) {};}}

\begin{scope}[xshift=0cm]
  \node[font=\sffamily\bfseries] at (0,5.75) {(a) Full fine-tuning};
  \tokens
  \stack{trainfill}{trainline}
  \node[align=center] at (0,1.65) {Transformer\\layers};
  \hidden
  \node[head] (ha) at (0,4.3) {Classification head};
  \node[small] at (0,5.15) {loss(prediction, label)};
  \draw[fwd] (-1.95,0) -- node[small, rotate=90, above] {forward} (-1.95,4.3);
  \draw[bwd] (1.95,5.0) -- node[small, rotate=-90, above] {backpropagation} (1.95,0.55);
  \node[small, font=\sffamily\footnotesize\bfseries] at (0,-0.95) {Trained: all weights};
\end{scope}

\begin{scope}[xshift=5.4cm]
  \node[font=\sffamily\bfseries] at (0,5.75) {(b) LoRA (parameter-efficient)};
  \tokens
  \stack{frozenfill}{frozenline}
  \node[align=center] at (-0.35,1.65) {Transformer\\layers};
  \foreach \y in {0.95,1.65,2.35} {
    \node[draw=trainline, fill=trainfill, minimum width=5mm, minimum height=3.5mm,
          inner sep=0pt, font=\sffamily\scriptsize] at (0.9,\y) {$BA$};}
  \node[head] (hb) at (0,4.3) {Classification head};
  \node[small] at (0,5.15) {loss(prediction, label)};
  \draw[fwd] (-1.95,0) -- node[small, rotate=90, above] {forward} (-1.95,4.3);
  \draw[bwd] (1.95,5.0) -- node[small, rotate=-90, above] {updates adapters + head} (1.95,0.55);
  \node[small, font=\sffamily\footnotesize\bfseries] at (0,-0.95) {Trained: low-rank adapters + head};
\end{scope}

\begin{scope}[xshift=10.8cm]
  \node[font=\sffamily\bfseries] at (0,5.75) {(c) Probing};
  \tokens
  \stack{frozenfill}{frozenline}
  \node[align=center] at (0,1.65) {Transformer\\layers};
  \hidden
  \node[small, draw=boxline, fill=white, rounded corners=2pt, inner sep=1.5pt] (pool) at (0,3.66)
       {masked mean pooling};
  \node[head] (hc) at (0,4.3) {MLP classifier};
  \node[small] at (0,5.15) {loss(prediction, label)};
  \draw[fwd] (-1.95,0) -- node[small, rotate=90, above] {forward} (-1.95,4.3);
  \draw[bwd] (1.95,5.0) -- node[small, rotate=-90, above, align=center] {backprop.} (1.95,3.95);
  \node[small, font=\sffamily\footnotesize\bfseries] at (0,-0.95) {Trained: MLP only};
\end{scope}

\begin{scope}[yshift=-1.7cm, xshift=1.2cm]
  \node[draw=trainline, fill=trainfill, minimum width=5mm, minimum height=3.5mm, inner sep=0pt] (l1) at (0,0) {};
  \node[small, right=1mm of l1] (t1) {trainable};
  \node[draw=frozenline, fill=frozenfill, minimum width=5mm, minimum height=3.5mm, inner sep=0pt,
        right=5mm of t1] (l2) {};
  \node[small, right=1mm of l2] (t2) {frozen};
  \draw[fwd] ([xshift=6mm]t2.east) -- ++(8mm,0) node[small, right] (t3) {forward pass};
  \draw[bwd] ([xshift=5mm]t3.east) -- ++(8mm,0) node[small, right] {gradient updates};
\end{scope}
\end{tikzpicture}}

%% file: sections/results.tex
\section{Results}
\begin{figure}[H]
  \centering
  \includegraphics[width=0.92\textwidth]{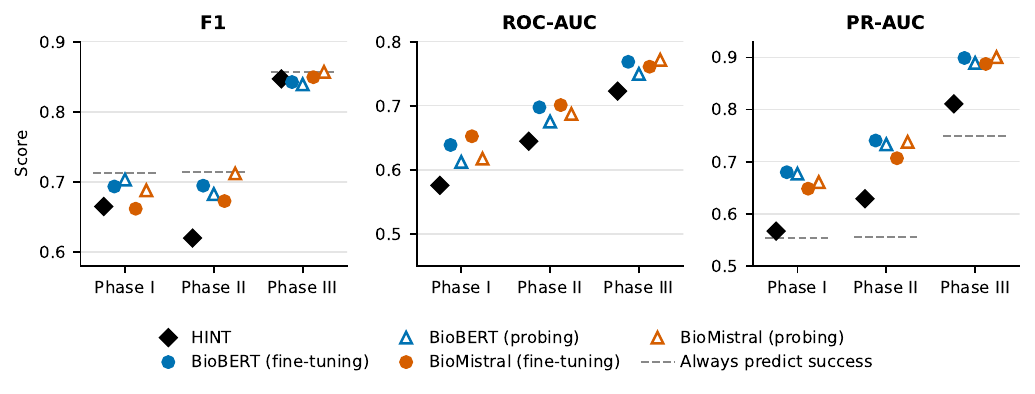}
  \caption{Test-set F1 (left), ROC-AUC (middle) and PR-AUC (right) by phase for HINT and the four language-model variants (values from Table~\ref{tab:phase-results}). Colour indicates the model and marker shape the training strategy. Dashed grey lines show the trivial classifier that predicts success for every trial (for ROC-AUC, chance level 0.5). Abbreviations: HINT, Hierarchical Interaction Network; PR-AUC, area under the precision--recall curve; ROC-AUC, area under the receiver operating characteristic curve.}
  \label{fig:performance-comparison}
\end{figure}

Table~\ref{tab:phase-results} and Fig.~\ref{fig:performance-comparison} report the test-set performance of the four language-model variants (BioBERT and BioMistral, each fine-tuned and probed) alongside the published HINT results and the trivial always-success reference.

\begin{table}[H]
\centering
\small
\setlength{\tabcolsep}{6pt}
\caption{Test-set performance by phase (I, II, III). $^\dagger$HINT results are taken from Fu et al.~\cite{fu2020hint}, where a separate model was trained for each phase on an earlier version of the TOP data; values are means over 30 runs (standard deviations 0.006--0.010). The last row is a trivial classifier that predicts success for every trial; its F1 and PR-AUC depend only on the test-set success rate (55.3\%, 55.5\% and 75.0\% for Phases~I--III).}
\label{tab:phase-results}
\begin{tabular}{lccccccccc}
\toprule
 & \multicolumn{3}{c}{F1 (phase)} & \multicolumn{3}{c}{ROC-AUC (phase)} & \multicolumn{3}{c}{PR-AUC (phase)} \\
\cmidrule(lr){2-4} \cmidrule(lr){5-7} \cmidrule(lr){8-10}
Model & I & II & III & I & II & III & I & II & III \\
\midrule
HINT$^\dagger$            & 0.665 & 0.620 & 0.847 & 0.576 & 0.645 & 0.723 & 0.567 & 0.629 & 0.811 \\
BioBERT (fine-tuning)     & 0.694 & 0.695 & 0.843 & 0.639 & 0.698 & 0.769 & 0.680 & 0.741 & 0.899 \\
BioMistral (fine-tuning)  & 0.662 & 0.673 & 0.850 & 0.653 & 0.701 & 0.761 & 0.649 & 0.707 & 0.888 \\
BioBERT (probing)         & 0.704 & 0.683 & 0.839 & 0.613 & 0.676 & 0.750 & 0.677 & 0.734 & 0.889 \\
BioMistral (probing)      & 0.688 & 0.712 & 0.857 & 0.618 & 0.688 & 0.772 & 0.661 & 0.738 & 0.901 \\
\midrule
Always predict success    & 0.713 & 0.714 & 0.857 & 0.500 & 0.500 & 0.500 & 0.553 & 0.555 & 0.750 \\
\bottomrule
\end{tabular}
\end{table}

All results come from a single training run per model. With 627 to 1,654 test trials per phase, the standard error of ROC-AUC is roughly 0.01--0.02, so many of the differences between the four language-model variants, which are typically below 0.03, are of the same order as this sampling uncertainty. The comparisons below should be read with this in mind.

\subsection{Comparison with HINT}
All four language-model variants achieved higher ROC-AUC and PR-AUC than HINT in every phase. Relative to HINT, ROC-AUC improved by 3.7--13.4\% and PR-AUC by 9.5--19.9\% (computed from the values in Table~\ref{tab:phase-results}). The largest gains were in Phase~I, where fine-tuned BioMistral reached a ROC-AUC of 0.653 (HINT: 0.576) and fine-tuned BioBERT a PR-AUC of 0.680 (HINT: 0.567).

F1 results were mixed, ranging from $-0.9\%$ to $+14.8\%$ relative to HINT. Three results were below HINT: fine-tuned BioMistral in Phase~I (0.662 vs.\ 0.665), and fine-tuned BioBERT (0.843) and BioBERT probing (0.839) in Phase~III (HINT: 0.847). Moreover, every model, including HINT, had an F1 at or below that of the trivial always-success classifier in every phase (0.713, 0.714 and 0.857). Because F1 at a fixed threshold is dominated by the proportion of successful trials on this benchmark, we treat the threshold-independent ROC-AUC and PR-AUC as the primary comparison. PR-AUC should likewise be read relative to its baseline: in Phase~III, for example, the highest PR-AUC (0.901) compares with 0.750 for the trivial classifier. Table~\ref{tab:normalised-prauc} expresses PR-AUC relative to this baseline. On this scale, HINT is barely above chance in Phase~I (0.030), whereas the language-model variants reach up to 0.283, 0.418 and 0.605 in Phases~I--III.

\begin{table}[H]
\centering
\small
\caption{Prevalence-normalised PR-AUC, $(\text{PR-AUC}-\pi)/(1-\pi)$, by phase, computed from Table~\ref{tab:phase-results} and the test-set success rates $\pi$ (55.3\%, 55.5\% and 75.0\%). A random classifier scores 0 and a perfect classifier 1. $^\dagger$HINT values are based on the published results (see Table~\ref{tab:phase-results}).}
\label{tab:normalised-prauc}
\begin{tabular}{lccc}
\toprule
Model & Phase I & Phase II & Phase III \\
\midrule
HINT$^\dagger$            & 0.030 & 0.166 & 0.245 \\
BioBERT (fine-tuning)     & 0.283 & 0.418 & 0.597 \\
BioMistral (fine-tuning)  & 0.214 & 0.342 & 0.553 \\
BioBERT (probing)         & 0.277 & 0.402 & 0.557 \\
BioMistral (probing)      & 0.241 & 0.411 & 0.605 \\
\bottomrule
\end{tabular}
\end{table}

The HINT results were obtained with per-phase training and different input modalities, and were not reproduced under our setting; the comparison is therefore indicative rather than fully controlled (Section~\ref{sec:limitations}).

\subsection{Performance across trial phases}
ROC-AUC, which does not depend on class balance, increased from Phase~I to Phase~III for every model, including HINT. For the language-model variants it ranged from 0.613--0.653 in Phase~I to 0.676--0.701 in Phase~II and 0.750--0.772 in Phase~III. F1 and PR-AUC were also highest in Phase~III, but part of this reflects the higher success rate in the Phase~III test set (75.0\%, compared with about 55\% in Phases~I and~II), which raises the baseline for both metrics. F1 did not increase consistently from Phase~I to Phase~II. Phase~I trials are also the smallest group in the training data (711 of 8,436 trials, or 8\%), so weaker Phase~I performance may partly reflect limited training data rather than a harder task.

\subsection{Fine-tuning}
Fine-tuned models achieved the highest ROC-AUC and PR-AUC in Phases~I and~II: fine-tuned BioMistral had the highest ROC-AUC (0.653 and 0.701), and fine-tuned BioBERT the highest PR-AUC (0.680 and 0.741). For BioBERT, fine-tuning exceeded probing on both ROC-AUC and PR-AUC in all three phases, although by small margins (0.003--0.026).

\begin{figure}[H]
  \centering
  \includegraphics[width=0.92\textwidth]{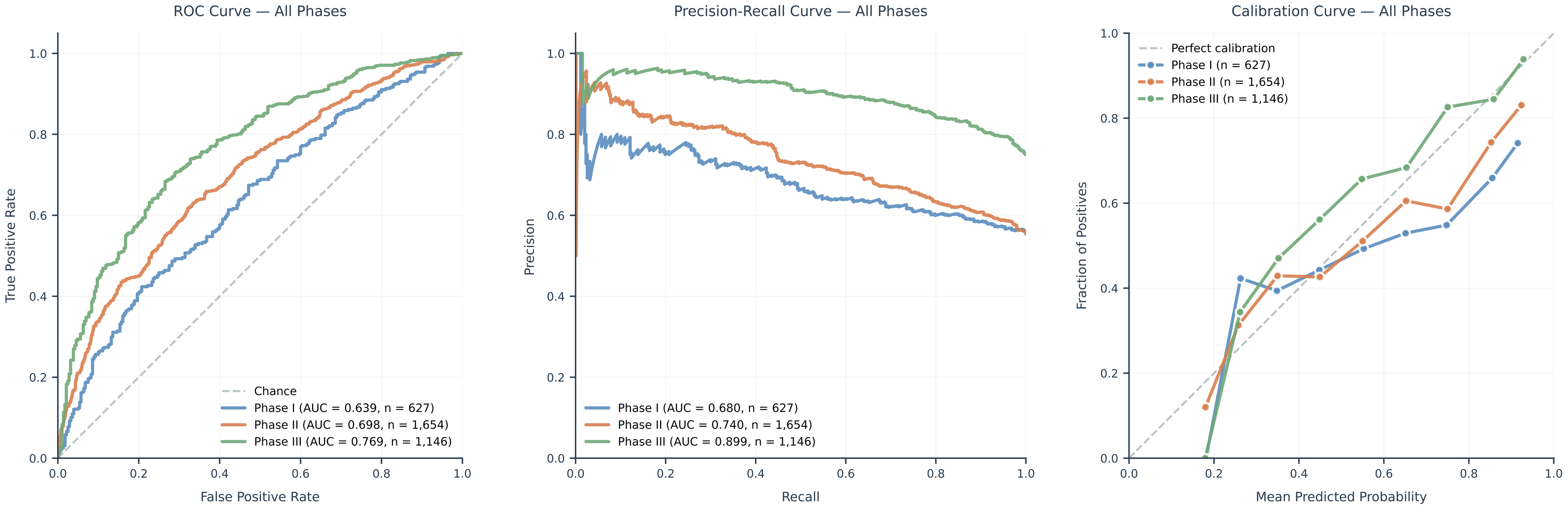}
  \caption{ROC curves (left), precision--recall curves (middle) and calibration curves (right) for fine-tuned BioBERT on the test set of each phase (Phase~I, $n=627$; Phase~II, $n=1{,}654$; Phase~III, $n=1{,}146$). The dashed lines show chance-level performance (ROC) and perfect calibration (calibration plot). Abbreviations: AUC, area under the curve; ROC, receiver operating characteristic.}
  \label{fig:biobert-finetuning-results}
\end{figure}

\begin{figure}[H]
  \centering
  \includegraphics[width=0.92\textwidth]{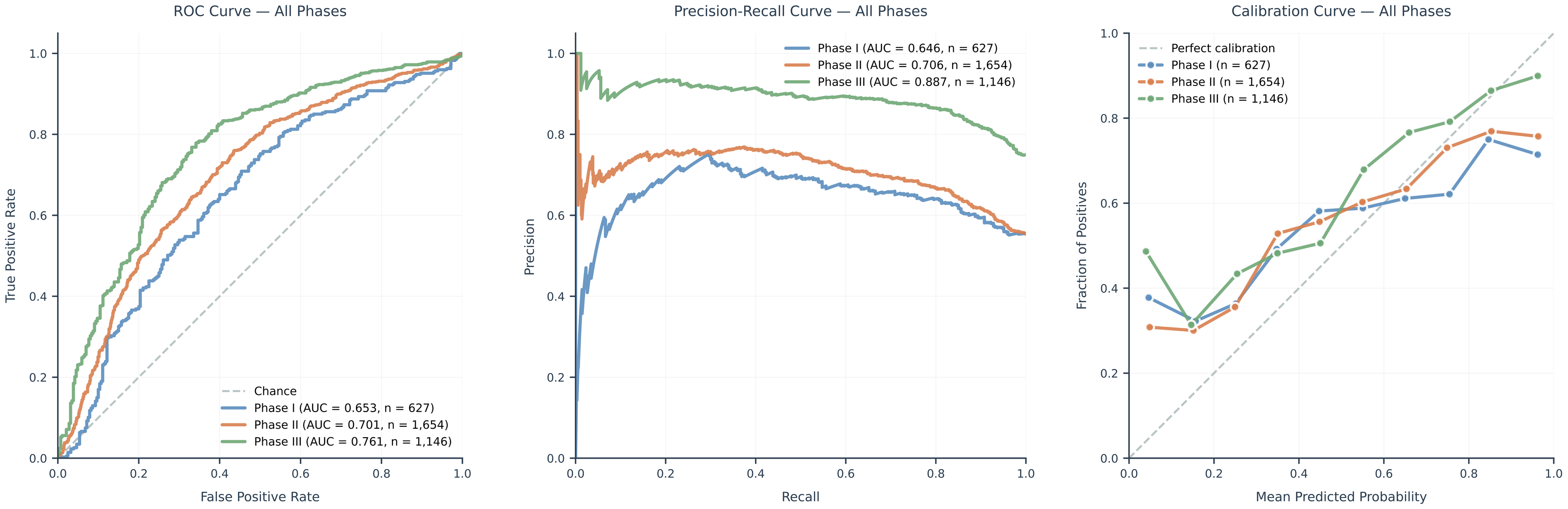}
  \caption{ROC curves (left), precision--recall curves (middle) and calibration curves (right) for BioMistral fine-tuned with LoRA on the test set of each phase (Phase~I, $n=627$; Phase~II, $n=1{,}654$; Phase~III, $n=1{,}146$). Dashed lines and abbreviations as in Fig.~\ref{fig:biobert-finetuning-results}; LoRA, low-rank adaptation.}
  \label{fig:biomistral-finetuning-results}
\end{figure}

The calibration curves (Figs.~\ref{fig:biobert-finetuning-results} and~\ref{fig:biomistral-finetuning-results}) show that fine-tuned BioBERT produced probabilities between about 0.2 and 0.9 that broadly follow the diagonal. Its Phase~III probabilities are somewhat too low in the middle of the range, and its Phase~I probabilities too high above about 0.6. Fine-tuned BioMistral produced a wider range of probabilities; it underestimated the probability of success below about 0.3, where the observed success rate was 0.3--0.5, and overestimated it above 0.8 in Phases~I and~II.

\subsection{Probing}
\begin{figure}[H]
  \centering
  \includegraphics[width=0.92\textwidth]{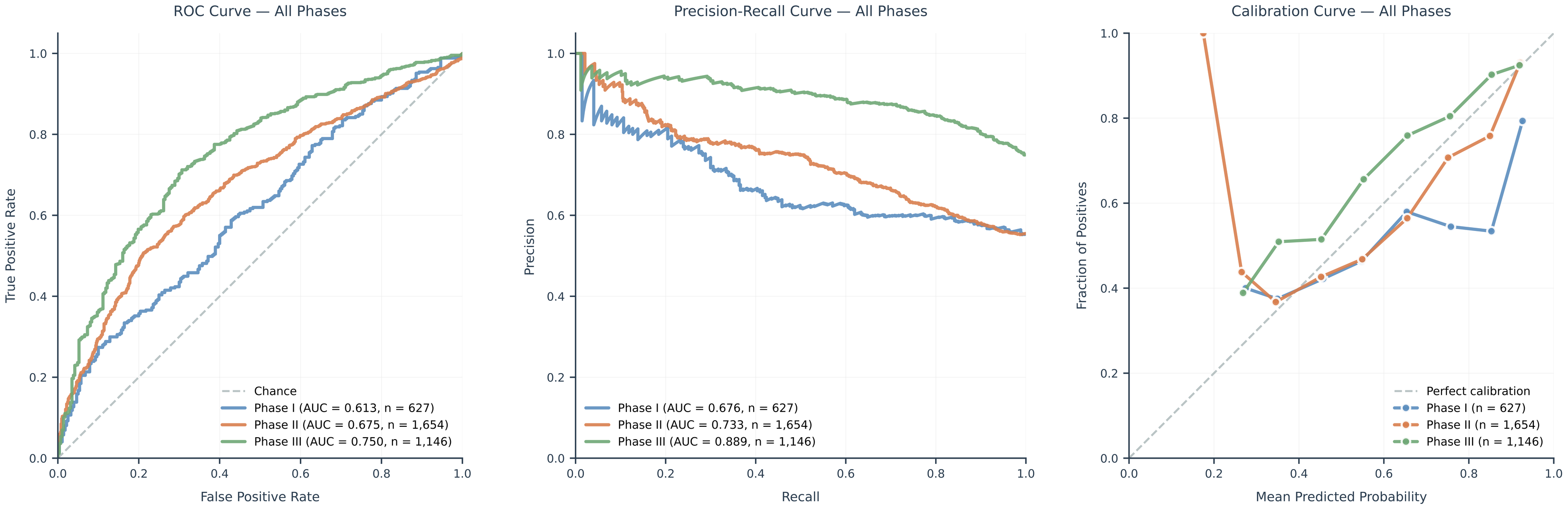}
  \caption{ROC curves (left), precision--recall curves (middle) and calibration curves (right) for BioBERT probing on the test set of each phase (Phase~I, $n=627$; Phase~II, $n=1{,}654$; Phase~III, $n=1{,}146$). Dashed lines and abbreviations as in Fig.~\ref{fig:biobert-finetuning-results}.}
  \label{fig:biobert-probing-results}
\end{figure}

\begin{figure}[H]
  \centering
  \includegraphics[width=0.92\textwidth]{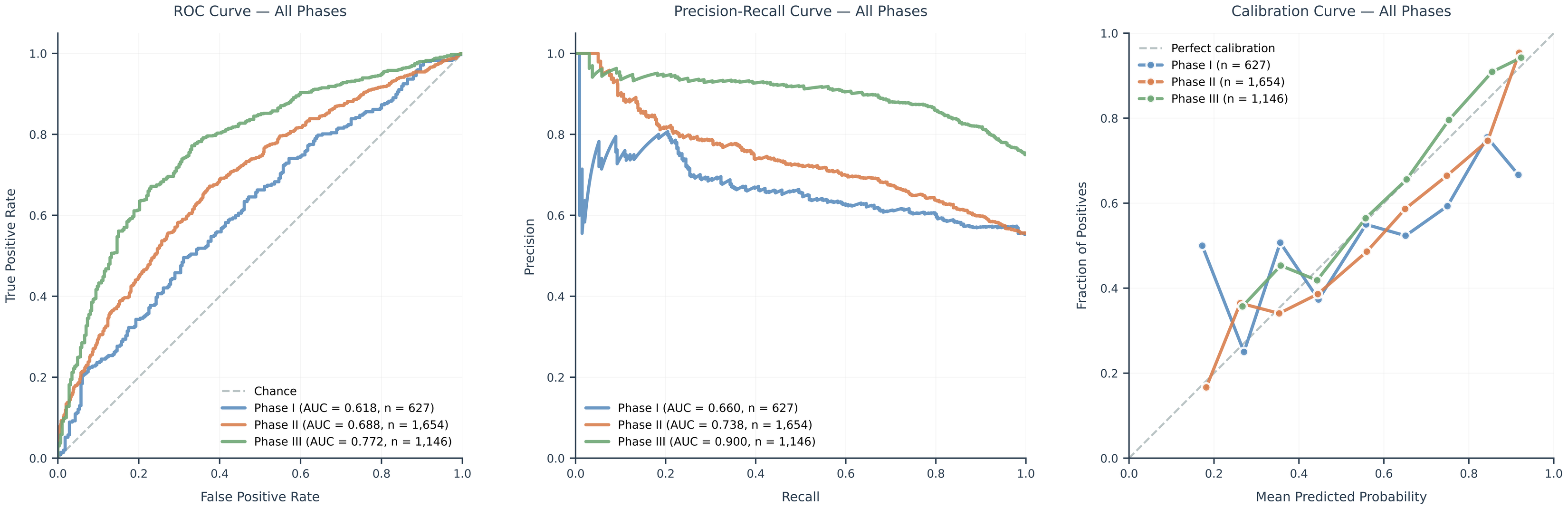}
  \caption{ROC curves (left), precision--recall curves (middle) and calibration curves (right) for BioMistral probing on the test set of each phase (Phase~I, $n=627$; Phase~II, $n=1{,}654$; Phase~III, $n=1{,}146$). Dashed lines and abbreviations as in Fig.~\ref{fig:biobert-finetuning-results}.}
  \label{fig:biomistral-probing-results}
\end{figure}

Probing frozen representations performed comparably to fine-tuning (Figs.~\ref{fig:biobert-probing-results} and~\ref{fig:biomistral-probing-results}). BioMistral probing achieved the highest Phase~III ROC-AUC (0.772) and PR-AUC (0.901) of all models. Compared with fine-tuned BioMistral, it had higher F1 in all three phases (0.688, 0.712 and 0.857 vs.\ 0.662, 0.673 and 0.850) and higher PR-AUC in all three phases (0.661, 0.738 and 0.901 vs.\ 0.649, 0.707 and 0.888), but lower ROC-AUC in Phases~I and~II (0.618 and 0.688 vs.\ 0.653 and 0.701). BioBERT probing was within 0.03 of fine-tuned BioBERT on every metric and had the highest Phase~I F1 (0.704), although this is still below the trivial reference of 0.713. Together, these results suggest that a substantial part of the predictive signal is already present in the pretrained biomedical representations. The comparison with LoRA should be read with caution, however: LoRA was run with a single configuration (learning rate $3\times10^{-5}$, mini-batch size 2, adapters on the query and value projections only), which may understate what fine-tuning can achieve.

In the calibration curves, BioMistral probing lies closest to the diagonal across phases, with more variability in Phase~I. BioBERT probing is less consistent: its Phase~III probabilities are too low in the middle of the range, its Phase~I probabilities too high above about 0.7, and its lowest-probability Phase~II bin shows an outlying observed rate of 1.0, most likely because it contains very few trials.

%% file: sections/discussion.tex
\section{Discussion}
Representing trials as text built from registry fields and processing that text with biomedical language models ranked trials more accurately than the published HINT results in all three phases, with relative PR-AUC gains of 10--20\%. This held for both model sizes and both training strategies, which suggests that the free-text and design fields of trial registrations carry predictive signal that structured interaction models do not fully exploit. The gains were clear for the threshold-independent metrics but not for F1, which varied little between models and did not clearly exceed a trivial classifier.

\textbf{Probing versus fine-tuning.}
The relative merit of probing depended on the model. For BioMistral, frozen embeddings with an MLP classifier matched or exceeded LoRA fine-tuning on F1 and PR-AUC in every phase and had the highest observed Phase~III ROC-AUC and PR-AUC. One interpretation is that a 7B model pretrained on biomedical literature already produces representations that separate successful from unsuccessful trials; another, at least as likely, is that our LoRA configuration did not reach its full potential, since it used a single setting with a learning rate of $3\times10^{-5}$, a mini-batch size of 2 and adapters on the query and value projections only. For BioBERT, full fine-tuning was slightly better than probing on ROC-AUC and PR-AUC in all phases; one possible explanation, which we did not test, is that a smaller model benefits more from task-specific adaptation. In both cases the differences were small. From a practical standpoint, probing requires one forward pass of the frozen model per field (ten per trial) to compute embeddings, which are computed once and stored, after which only a small classifier is trained (1.2M parameters for BioBERT and 2.9M for BioMistral), whereas full fine-tuning updates all 108M BioBERT parameters and LoRA trains 13.6M adapter parameters (0.19\% of BioMistral) but still requires forward and backward passes through the full 7B model at every step.

\textbf{Relation to prior work.}
Other methods have also reported improvements over HINT on this task. LINT, which likewise relies on trial text, reported higher ROC-AUC values than ours in Phases~I and~II, but on trials of biologic interventions only~\cite{gao2024lint}. HINT with selective classification reached a Phase~III PR-AUC of 0.902 on the subset of trials it chose to classify~\cite{lu2024uncertainty}, and SPOT and LIFTED report gains over HINT through meta-learning and mixture-of-experts architectures~\cite{wang2023spot,zheng2025lifted}. Because these studies differ in data subsets, splits or evaluation protocols, a comparison on identical test sets would be needed to rank the approaches.

\textbf{Choice of metrics.}
On the TOP benchmark, F1 at a fixed threshold is strongly determined by class prevalence: a classifier that predicts success for every trial reaches F1 values of 0.713, 0.714 and 0.857 in Phases~I--III, which none of the evaluated models, including HINT, clearly exceeded. This applies equally to earlier work that reports F1 on this benchmark. We therefore recommend that studies on TOP report the prevalence baseline alongside F1 and PR-AUC (or a prevalence-normalised PR-AUC), emphasise threshold-independent metrics, and, where a decision threshold is needed, choose it on the validation set and complement F1 with metrics such as balanced accuracy or the Matthews correlation coefficient.

\subsection{Limitations}\label{sec:limitations}
\textbf{Statistical uncertainty.} Each model was trained once, and no confidence intervals or significance tests are reported. Many differences between the language-model variants are of the same order as the sampling uncertainty of the test sets.

\textbf{Baseline comparability.} HINT results were taken from the original publication rather than reproduced. HINT was trained separately for each phase with different input modalities, whereas our models were trained on pooled data from all phases. We also did not compare against general-domain models, such as term frequency--inverse document frequency (TF-IDF) features with logistic regression, BERT-base or Mistral-7B-Instruct-v0.1, so the improvement cannot be attributed specifically to biomedical pretraining.

\textbf{Registry snapshot.} Trial records were downloaded in September 2024 and reflect their latest version rather than the version available when each trial started, so the text may include edits made after a trial began.

\textbf{Pretraining overlap.} BioBERT and BioMistral were pretrained on PubMed and PubMed Central, which include publications reporting the results of some test-set trials. The models may therefore partly recall reported outcomes rather than infer them from trial design.

\textbf{Scope.} The study uses a single benchmark of small-molecule trials, English-language registry text and a binary outcome label.

Future work will address these limitations and explore combining textual representations with the structured drug and disease features used by HINT.

%% file: sections/conclusion.tex
\section{Conclusion}
We evaluated BioBERT and BioMistral for phase-level clinical trial outcome prediction on the TOP benchmark, using a text-only representation of each trial built from ten ClinicalTrials.gov fields. Both fine-tuned and probed models ranked trials more accurately than the published HINT results in all three phases, as measured by ROC-AUC and PR-AUC, while F1 remained close to that of a trivial classifier for all models. Under the configurations tested, probing frozen representations performed comparably to fine-tuning, and BioMistral probing had the highest observed Phase~III ROC-AUC and PR-AUC. This suggests that pretrained biomedical language models already encode much of the relevant signal, and that probing offers a computationally cheaper way to use them: the frozen model is run once to compute embeddings, and only a small classifier is trained. Confidence intervals, a simple text baseline, a controlled re-evaluation of HINT and the use of originally registered trial records are the most important next steps for confirming these findings.

%% file: sections/declarations.tex
\section*{Acknowledgements}
We thank Ahmed Cheikh Rouhou and Imen Jendoubi for reviewing the manuscript and for their helpful comments, which improved its quality.

\section*{Data availability}
The TOP benchmark is publicly available from its GitHub repository~\cite{futianfan_top_data} under a non-commercial licence; we used the data files as of commit \texttt{d1f5c4a}. Trial records were obtained from the public ClinicalTrials.gov registry~\cite{clinicaltrialsgov}.





%% file: references.bib
@article{sun2022why90,
  author  = {Sun, Duxin and Gao, Wei and Hu, Hongxiang and Zhou, Simon},
  title   = {Why 90\% of clinical drug development fails and how to improve it?},
  journal = {Acta Pharmaceutica Sinica B},
  year    = {2022},
  volume  = {12},
  number  = {7},
  pages   = {3049--3062},
  doi     = {10.1016/j.apsb.2022.02.002}
}

@article{fogel2018factors,
  author  = {Fogel, David B.},
  title   = {Factors associated with clinical trials that fail and opportunities for improving the likelihood of success: A review},
  journal = {Contemporary Clinical Trials Communications},
  year    = {2018},
  volume  = {11},
  pages   = {156--164},
  doi     = {10.1016/j.conctc.2018.08.001}
}

@article{wong2019estimation,
  author  = {Wong, Chi Heem and Siah, Kien Wei and Lo, Andrew W.},
  title   = {Estimation of clinical trial success rates and related parameters},
  journal = {Biostatistics},
  year    = {2019},
  volume  = {20},
  number  = {2},
  pages   = {273--286},
  doi     = {10.1093/biostatistics/kxx069}
}

@article{gayvert2016data,
  author  = {Gayvert, Kaitlyn M. and Madhukar, Neel S. and Elemento, Olivier},
  title   = {A data-driven approach to predicting successes and failures of clinical trials},
  journal = {Cell Chemical Biology},
  year    = {2016},
  volume  = {23},
  number  = {10},
  pages   = {1294--1301},
  doi     = {10.1016/j.chembiol.2016.07.023}
}

@article{lo2019machine,
  author  = {Lo, Andrew W. and Siah, Kien Wei and Wong, Chi Heem},
  title   = {Machine learning with statistical imputation for predicting drug approvals},
  journal = {Harvard Data Science Review},
  year    = {2019},
  volume  = {1},
  number  = {1},
  doi     = {10.1162/99608f92.5c5f0525}
}

@misc{hong2020ensemble,
  author       = {Hong, Zhen-Yu and Shim, Jooyong and Son, Woo Chan and Hwang, Changha},
  title        = {Predicting successes and failures of clinical trials with an ensemble {LS-SVR}},
  howpublished = {medRxiv preprint},
  year         = {2020},
  doi          = {10.1101/2020.02.05.20020636}
}

@article{fu2020hint,
  author  = {Fu, Tianfan and Huang, Kexin and Xiao, Cao and Glass, Lucas M. and Sun, Jimeng},
  title   = {{HINT}: Hierarchical interaction network for clinical-trial-outcome predictions},
  journal = {Patterns},
  year    = {2022},
  volume  = {3},
  number  = {4},
  pages   = {100445},
  doi     = {10.1016/j.patter.2022.100445}
}

@misc{futianfan_top_data,
  author       = {Fu, Tianfan},
  title        = {Clinical-trial-outcome-prediction: {TOP} benchmark dataset},
  howpublished = {GitHub repository},
  url          = {https://github.com/futianfan/clinical-trial-outcome-prediction},
  year         = {2021},
  note         = {Data files at commit d1f5c4a (30 December 2021); released for non-commercial use only}
}

@misc{clinicaltrialsgov,
  author       = {{U.S. National Library of Medicine}},
  title        = {{ClinicalTrials.gov}},
  url          = {https://clinicaltrials.gov},
  year         = {2024},
  note         = {Records downloaded 12 September 2024}
}

@article{lee2020biobert,
  author  = {Lee, Jinhyuk and Yoon, Wonjin and Kim, Sungdong and Kim, Donghyeon and Kim, Sunkyu and So, Chan Ho and Kang, Jaewoo},
  title   = {{BioBERT}: a pre-trained biomedical language representation model for biomedical text mining},
  journal = {Bioinformatics},
  year    = {2020},
  volume  = {36},
  number  = {4},
  pages   = {1234--1240},
  doi     = {10.1093/bioinformatics/btz682}
}

@inproceedings{labrak2024biomistral,
  author    = {Labrak, Yanis and Bazoge, Adrien and Morin, Emmanuel and Gourraud, Pierre-Antoine and Rouvier, Mickael and Dufour, Richard},
  title     = {{BioMistral}: A collection of open-source pretrained large language models for medical domains},
  booktitle = {Findings of the Association for Computational Linguistics: ACL 2024},
  year      = {2024},
  pages     = {5848--5864},
  doi       = {10.18653/v1/2024.findings-acl.348}
}

@misc{jiang2023mistral,
  author       = {Jiang, Albert Q. and Sablayrolles, Alexandre and Mensch, Arthur and Bamford, Chris and Chaplot, Devendra Singh and de las Casas, Diego and Bressand, Florian and Lengyel, Gianna and Lample, Guillaume and Saulnier, Lucile and Renard Lavaud, L{\'e}lio and Lachaux, Marie-Anne and Stock, Pierre and Le Scao, Teven and Lavril, Thibaut and Wang, Thomas and Lacroix, Timoth{\'e}e and El Sayed, William},
  title        = {Mistral {7B}},
  howpublished = {arXiv preprint arXiv:2310.06825},
  year         = {2023},
  doi          = {10.48550/arXiv.2310.06825}
}

@article{gu2021domain,
  author  = {Gu, Yu and Tinn, Robert and Cheng, Hao and Lucas, Michael and Usuyama, Naoto and Liu, Xiaodong and Naumann, Tristan and Gao, Jianfeng and Poon, Hoifung},
  title   = {Domain-specific language model pretraining for biomedical natural language processing},
  journal = {ACM Transactions on Computing for Healthcare},
  year    = {2021},
  volume  = {3},
  number  = {1},
  pages   = {1--23},
  doi     = {10.1145/3458754}
}

@article{singhal2023large,
  author  = {Singhal, Karan and Azizi, Shekoofeh and Tu, Tao and Mahdavi, S. Sara and Wei, Jason and Chung, Hyung Won and Scales, Nathan and Tanwani, Ajay and Cole-Lewis, Heather and Pfohl, Stephen and Payne, Perry and Seneviratne, Martin and Gamble, Paul and Kelly, Chris and Babiker, Abubakr and Sch{\"a}rli, Nathanael and Chowdhery, Aakanksha and Mansfield, Philip and Demner-Fushman, Dina and Ag{\"u}era y Arcas, Blaise and Webster, Dale and Corrado, Greg S. and Matias, Yossi and Chou, Katherine and Gottweis, Juraj and Tomasev, Nenad and Liu, Yun and Rajkomar, Alvin and Barral, Joelle and Semturs, Christopher and Karthikesalingam, Alan and Natarajan, Vivek},
  title   = {Large language models encode clinical knowledge},
  journal = {Nature},
  year    = {2023},
  volume  = {620},
  pages   = {172--180},
  doi     = {10.1038/s41586-023-06291-2}
}

@article{rasmy2021medbert,
  author  = {Rasmy, Laila and Xiang, Yang and Xie, Ziqian and Tao, Cui and Zhi, Degui},
  title   = {{Med-BERT}: pretrained contextualized embeddings on large-scale structured electronic health records for disease prediction},
  journal = {npj Digital Medicine},
  year    = {2021},
  volume  = {4},
  pages   = {86},
  doi     = {10.1038/s41746-021-00455-y}
}

@misc{gao2024lint,
  author       = {Gao, Chufan and Fu, Tianfan and Sun, Jimeng},
  title        = {Language interaction network for clinical trial approval estimation},
  howpublished = {arXiv preprint arXiv:2405.06662},
  year         = {2024},
  doi          = {10.48550/arXiv.2405.06662}
}

@inproceedings{zheng2025lifted,
  author    = {Zheng, Wenhao and Wang, Liaoyaqi and Peng, Dongshen and Xu, Hongxia and Li, Yun and Zhu, Hongtu and Fu, Tianfan and Yao, Huaxiu},
  title     = {{LIFTED}: Multimodal clinical trial outcome prediction via large language models and mixture-of-experts},
  booktitle = {Findings of the Association for Computational Linguistics: EMNLP 2025},
  year      = {2025},
  pages     = {7503--7517},
  doi       = {10.18653/v1/2025.findings-emnlp.396}
}

@article{lu2024uncertainty,
  author  = {Lu, Yingzhou and Chen, Tianyi and Hao, Nan and Van Rechem, Capucine and Chen, Jintai and Fu, Tianfan},
  title   = {Uncertainty quantification and interpretability for clinical trial approval prediction},
  journal = {Health Data Science},
  year    = {2024},
  volume  = {4},
  pages   = {0126},
  doi     = {10.34133/hds.0126}
}

@inproceedings{wang2023spot,
  author    = {Wang, Zifeng and Xiao, Cao and Sun, Jimeng},
  title     = {{SPOT}: Sequential predictive modeling of clinical trial outcome with meta-learning},
  booktitle = {Proceedings of the 14th ACM International Conference on Bioinformatics, Computational Biology, and Health Informatics},
  year      = {2023},
  pages     = {1--11}
}

@inproceedings{yue2024trialenroll,
  author    = {Yue, Ling and Xing, Sixue and Chen, Jintai and Fu, Tianfan},
  title     = {{TrialEnroll}: Predicting clinical trial enrollment success with deep \& cross network and large language models},
  booktitle = {Proceedings of the 15th ACM International Conference on Bioinformatics, Computational Biology and Health Informatics},
  year      = {2024},
  pages     = {1--9},
  doi       = {10.1145/3698587.3701375}
}

@article{teodoro2025scoping,
  author  = {Teodoro, Douglas and Naderi, Nona and Yazdani, Anthony and Zhang, Boya and Bornet, Alban},
  title   = {A scoping review of artificial intelligence applications in clinical trial risk assessment},
  journal = {npj Digital Medicine},
  year    = {2025},
  volume  = {8},
  pages   = {486},
  doi     = {10.1038/s41746-025-01886-7}
}

@article{belinkov2022probing,
  author  = {Belinkov, Yonatan},
  title   = {Probing classifiers: Promises, shortcomings, and advances},
  journal = {Computational Linguistics},
  year    = {2022},
  volume  = {48},
  number  = {1},
  pages   = {207--219},
  doi     = {10.1162/coli_a_00422}
}

@inproceedings{hu2021lora,
  author    = {Hu, Edward J. and Shen, Yelong and Wallis, Phillip and Allen-Zhu, Zeyuan and Li, Yuanzhi and Wang, Shean and Wang, Lu and Chen, Weizhu},
  title     = {{LoRA}: Low-rank adaptation of large language models},
  booktitle = {International Conference on Learning Representations (ICLR)},
  year      = {2022}
}

@inproceedings{devlin2019bert,
  author    = {Devlin, Jacob and Chang, Ming-Wei and Lee, Kenton and Toutanova, Kristina},
  title     = {{BERT}: Pre-training of deep bidirectional transformers for language understanding},
  booktitle = {Proceedings of the 2019 Conference of the North American Chapter of the Association for Computational Linguistics: Human Language Technologies (NAACL-HLT)},
  year      = {2019},
  pages     = {4171--4186},
  doi       = {10.18653/v1/N19-1423}
}

@article{huang2020deeppurpose,
  author  = {Huang, Kexin and Fu, Tianfan and Glass, Lucas M. and Zitnik, Marinka and Xiao, Cao and Sun, Jimeng},
  title   = {{DeepPurpose}: a deep learning library for drug--target interaction prediction},
  journal = {Bioinformatics},
  year    = {2020},
  volume  = {36},
  number  = {22--23},
  pages   = {5545--5547},
  doi     = {10.1093/bioinformatics/btaa1005}
}

@inproceedings{choi2017gram,
  author    = {Choi, Edward and Bahadori, Mohammad Taha and Song, Le and Stewart, Walter F. and Sun, Jimeng},
  title     = {{GRAM}: Graph-based attention model for healthcare representation learning},
  booktitle = {Proceedings of the 23rd ACM SIGKDD International Conference on Knowledge Discovery and Data Mining},
  year      = {2017},
  pages     = {787--795},
  doi       = {10.1145/3097983.3098126}
}

@inproceedings{kingma2015adam,
  author    = {Kingma, Diederik P. and Ba, Jimmy},
  title     = {Adam: A method for stochastic optimization},
  booktitle = {International Conference on Learning Representations (ICLR)},
  year      = {2015}
}
